\documentclass[conference,a4paper]{APSIPA2021}
\usepackage{multirow}
\usepackage{graphicx}
\usepackage{amsmath}
\usepackage[psamsfonts]{amssymb}
\usepackage{amsxtra}
\usepackage{threeparttable}
\graphicspath{{figs/}}
\usepackage{booktabs}
\usepackage{etoolbox}
\usepackage{siunitx}
\usepackage{cite}
\usepackage{algorithm,algorithmic}
\usepackage[caption=false, font=footnotesize]{subfig} 

\newcommand*\rot{\rotatebox{90}}

\begin{document}

\title{
A Protection Method of Trained CNN Model Using Feature Maps Transformed With Secret Key From Unauthorized Access
}

\author{%
\authorblockN{%
MaungMaung AprilPyone and
Hitoshi Kiya
}
\authorblockA{%
Tokyo Metropolitan University, Tokyo, Japan
}
%
%
}

\maketitle
\thispagestyle{empty}

\begin{abstract}
In this paper, we propose a model protection method for convolutional neural networks (CNNs) with a secret key so that authorized users get a high classification accuracy, and unauthorized users get a low classification accuracy.
The proposed method applies a block-wise transformation with a secret key to feature maps in the network.
Conventional key-based model protection methods cannot maintain a high accuracy when a large key space is selected.
In contrast, the proposed method not only maintains almost the same accuracy as non-protected accuracy, but also has a larger key space.
Experiments were carried out on the CIFAR-10 dataset, and results show that the proposed model protection method outperformed the previous key-based model protection methods in terms of classification accuracy, key space, and robustness against key estimation attacks and fine-tuning attacks.
\end{abstract}

\section{Introduction}
Convolutional neural networks (CNNs) are a type of deep neural network (DNN) inspired by the human visual system and are ubiquitous in the field of computer vision.
Recent advances in deep learning show that CNNs have lead to major breakthroughs due to its effectiveness and efficiency~\cite{Lecun2015}.
Current commercial applications in image recognition, object detection and semantic segmentation are primarily powered by CNNs.
Therefore, CNNs have become \emph{de facto} standards for visual recognition systems in many different applications.

However, training successful CNNs requires three ingredients: huge amount of data, GPU-accelerated computing resources and efficient algorithms, and is not a trivial task.
For example, the dataset for ImageNet Large Scale Visual Recognition Challenge (ILSVRC 2012) contains about 1.28 million images, and training on such a dataset takes days and weeks even on GPU-accelerated machines. In fact, collecting images and labeling them are also costly, and will also consume a massive amount of resources. Moreover, algorithms used in training a CNN model may be patented or have restricted licenses. Therefore, trained CNNs have great business value. Considering the expenses necessary for the expertise, money, and time taken to train a CNN model, a model should be regarded as a kind of intellectual property (IP).

There are two aspects of IP protection for DNN models: ownership verification and access control.
The former focuses on identifying the ownership of the models and the latter addresses protecting the functionality of DNN models from unauthorized access.
Ownership verification methods are inspired by digital watermarking, and embed watermarks into DNN models so that the embedded watermarks are used to verify the ownership of the models in question~\cite{2017-ICMR-Uchida, 2018-USENIX-Yossi,2018-ACCCS-Zhang,2018-Arxiv-Rouhani,2019-NIPS-Fan, 2019-MIPR-Sakazawa, 2020-NCA-Le, maung2021piracy}.
Although the above watermarking methods can facilitate the identification of the ownership of the models, in reality, a stolen model can be exploited in many different ways.
For example, an attacker can use a model for their own benefit without arousing suspicion, or a stolen model can be used for model inversion attacks~\cite{2015-CCCS-Fredrikson} and adversarial attacks~\cite{2014-ICLR-Szegedy}.
Therefore, it is crucial to investigate mechanisms to protect DNN models from authorized access and misuse.
In this paper, we focus on protecting a model from misuse when it has been stolen (i.e., access control).

A method to model protection against unauthorized access was inspired by adversarial examples and proposed to utilize secret perturbation to control the access of a model~\cite{chen2018protect}.
Another study introduced a secret key to protect a model~\cite{pyone2020training, maungmaung2021protection}.
The secret key-based protection method~\cite{maungmaung2021protection} uses a key-based transformation which was originally used by an adversarial defense in~\cite{2020-Arxiv-Maung}, which was in turn inspired by perceptual image encryption methods~\cite{kawamura2020privacy,2018-ICCETW-Tanaka, 2019-Access-Warit, 2019-TIFS-Chuman, 2019-ICIP-Warit, 2019-APSIPAT-Warit, 2017-IEICE-Kurihara, chuman2018security}.
This model protection method utilizes a secret key in such a way that a stolen model cannot be used to its full capacity without a correct secret key.
However, in the previous model protection method~\cite{maungmaung2021protection}, when a large block size is used, the accuracy drops, and when a small block size is used, key estimation attacks are possible due to the relatively small key space.

Therefore, for the first time, in this paper, we propose a model protection method by applying key-based transformation to feature maps.
The proposed model protection method not only achieves a high classification accuracy (i.e., almost the same as non-protected accuracy), but also increases the key space substantially.
We make the following contributions in this paper.
\begin{itemize}
\item We propose a model protection method with a secret key which improves the classification accuracy and increases the key space.
\item We conduct relevant attacks to verify the effectiveness of the proposed model protection method.
\end{itemize}
In experiments, the proposed model protection method is confirmed to outperform the previous key-based model protection methods.

\section{Related Work}
\subsection{Ownership Verification}
Ownership verification is a concept to protect intellectual property of DNN models, in which digital watermarking techniques are adopted to embed watermarks into DNN models like copyright protection of media contents.
The ownership is verified by using the extracted watermark to detect the intellectual property of copyright infringement.

There are mainly two approaches in DNN model watermarking: white-box and black-box.
In white-box methods, a watermark is embedded to model weights by an embedding regularizer during training.
Therefore, the access to the model weights is required for extracting the watermark embedded in the model as in~\cite{2017-ICMR-Uchida, 2018-Arxiv-Chen, 2018-Arxiv-Rouhani, 2019-NIPS-Fan}.
In black-box model watermarking methods, an inspector observes the input and output of a model in doubt to verify the ownership as in~\cite{2018-USENIX-Yossi,2018-ACCCS-Zhang,2019-NIPS-Fan,2019-MIPR-Sakazawa,2020-NCA-Le,maung2021piracy}.
Thus, the access to the model weights is not required to verify the ownership in the black-box approaches.

Model watermarking methods in general focus on identifying the ownership only when the model is in question.
The functionality of the model is not protected regardless of the ownership.
Therefore, methods for protecting DNN models from unauthorized use are put forward beyond the ownership verification.

\subsection{Access Control}
One straightforward way of protecting a model from illicit use is to encrypt the trained model weights by the traditional cryptographic methods such as advanced encryption standard (AES).
In this case, to be able to use the protected model, rightful users have to decrypt the model.
There are millions of parameters in modern DNN models, so encrypting/decrypting all the parameters is computationally expensive under the traditional cryptography in general.
Besides, once the model is decrypted, it becomes vulnerable for IP thefts.
Therefore, researchers have proposed to embed a key to the model's structure by other means.
In the literature, there are two prior methods for protecting DNN models against unauthorized access.

The first method~\cite{chen2018protect} utilizes an anti-piracy transform module which is a secret perturbation network in such a way that the secret perturbation is crucial to the model's decision.
In other words, only the rightful users who have access to the secret perturbation can use the model properly.
However, this method~\cite{chen2018protect} requires to training a perturbation network along with the classification network so that the optimal perturbation can be learned.
In addition, the classification accuracy of the method~\cite{chen2018protect} slightly drops compared to non-protected models under the same training settings.

The second method~\cite{maungmaung2021protection} adopts a block-wise transformation with a secret key from an adversarial defense~\cite{2020-Arxiv-Maung} to protect the model against unauthorized access.
Images are transformed by a block-wise transformation with a secret key and transformed images are used for training and testing a model~\cite{maungmaung2021protection}.
When using a large block size, the second method drops the classification accuracy, and key estimation attacks are possible when using a small block size.
Therefore, in this paper, we aim to improve these issues on the classification accuracy and key space.

\section{Proposed Model Protection Method}
\subsection{Overview}
\begin{figure*}[t]
\centering
\subfloat[]{\includegraphics[width=\linewidth]{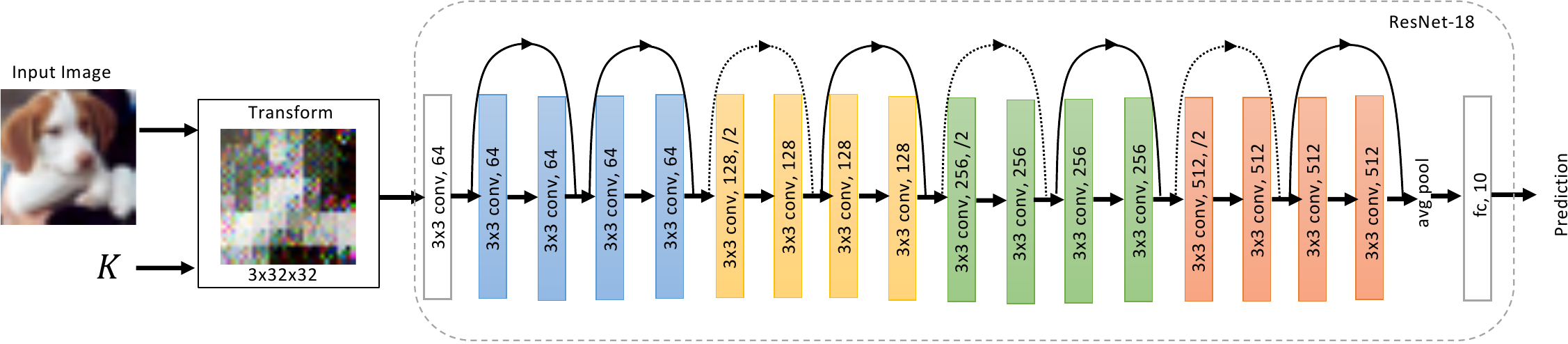}%
\label{fig:protection1}}
\hfil
\subfloat[]{\includegraphics[width=\linewidth]{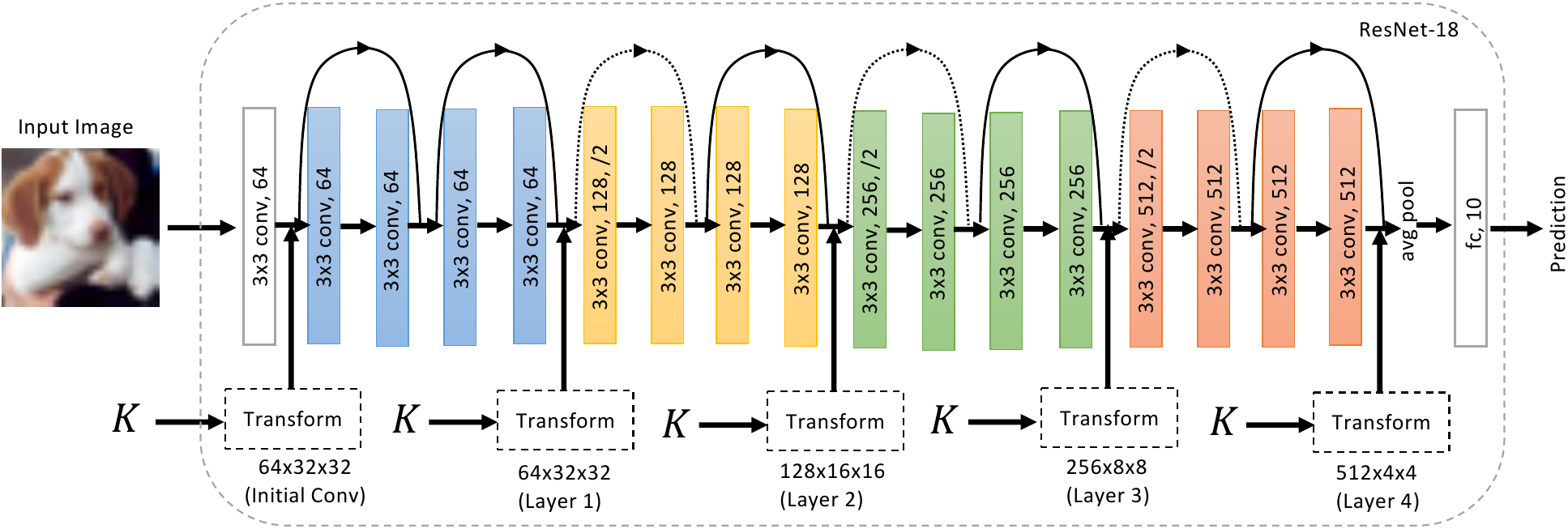}%
\label{fig:protection2}}
\caption{Prediction pipeline of model protection methods for ResNet-18. (a) Model protection with transformed images~\cite{maungmaung2021protection}. (b) Proposed model protection with transformed feature maps.\label{fig:overview}}
\end{figure*}

In the previous key-based model protection~\cite{maungmaung2021protection}, input images are transformed by a block-wise transformation with secret key $K$ prior to training and testing a model.
Key $K$ belongs to the model owner, and it needs to be stored at the model deployment (e.g., service provider).
Figure~\ref{fig:protection1} depicts the prediction pipeline of the model with the previous key-based model protection~\cite{maungmaung2021protection}.
In contrast, instead of transforming an input image, one or more feature maps in the network are transformed by a block-wise transformation with secret key $K$ in the proposed model protection (Fig.~\ref{fig:protection2}).
In the figure, the ResNet-18 is used as an example for illustrating the proposed method, and it can be replaced with other CNN architectures.

\begin{figure*}[t]
\centering
\subfloat[]{\includegraphics[width=0.5\linewidth]{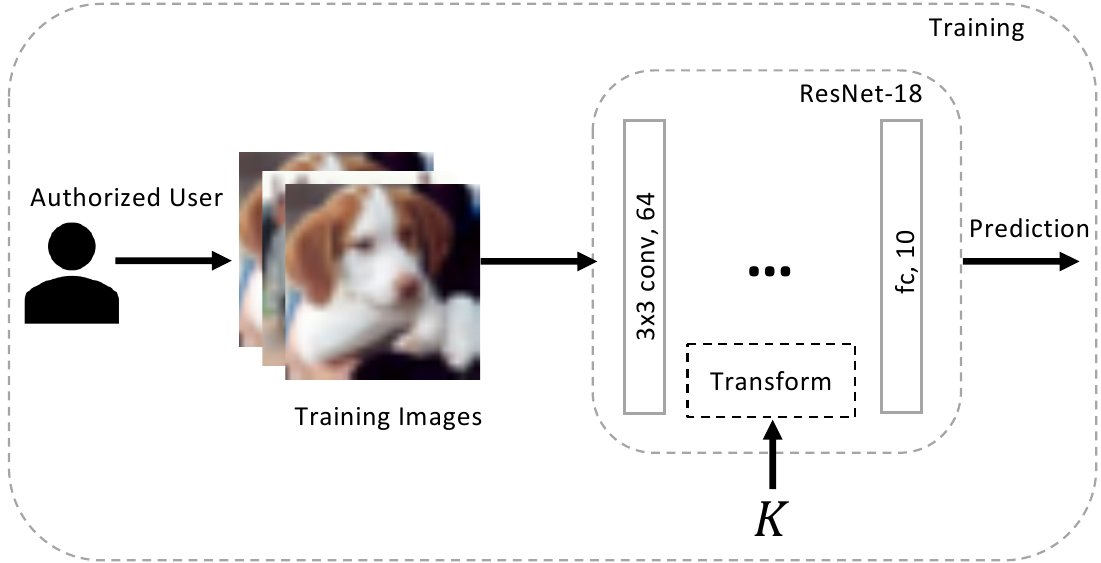}%
\label{fig:training}}
\hfil
\subfloat[]{\includegraphics[width=0.5\linewidth]{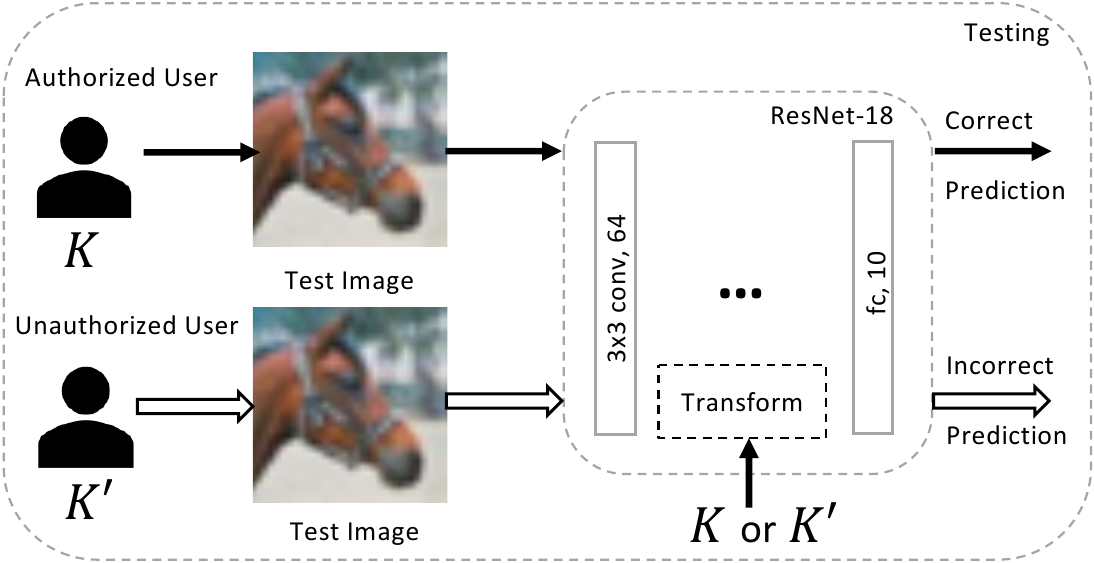}%
\label{fig:testing}}
\caption{Scenario of the proposed model protection. (a) Training process. (b) Testing process.\label{fig:framework}}
\end{figure*}

An overview of the proposed model protection is depicted in Fig.~\ref{fig:framework}.
A block-wise transformation with a secret key is applied to feature maps in the network (e.g., ResNet-18), where the modified network is trained by using secret key $K$ as shown in Fig.~\ref{fig:training}.
The model predicts a test image correctly only for authorized users with secret key $K$, and the model provides incorrect predictions for unauthorized users with incorrect key $K'$ (Fig.~\ref{fig:testing}).
Accordingly, a stolen model cannot be used to its full capacity when secret key $K$ is not available.

\subsection{Block-Wise Transformation with Secret Key for Feature Maps}
\begin{figure*}[t]
\begin{center}
\includegraphics[width=\linewidth]{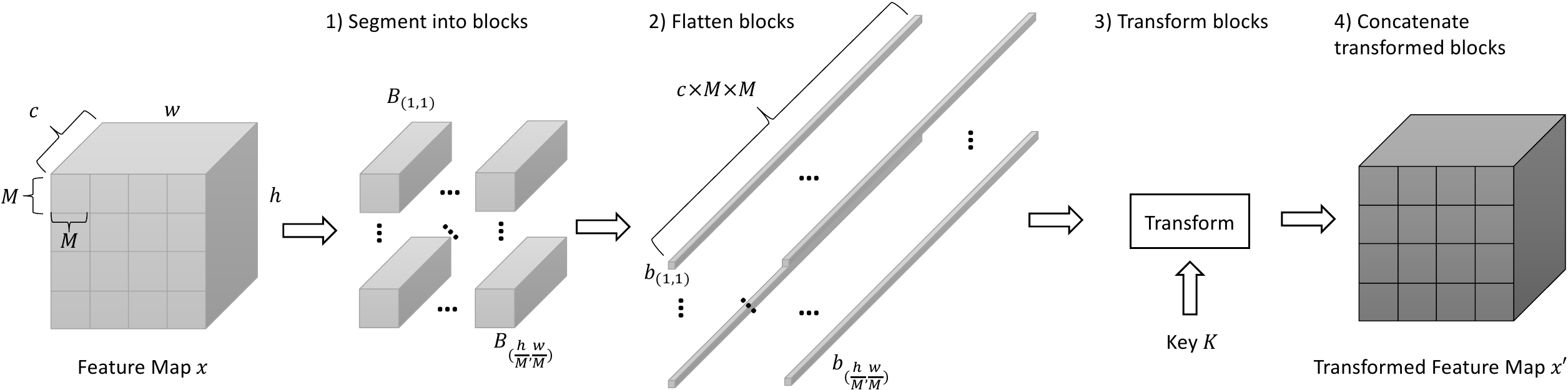}
\end{center}
\caption{Process of transforming a feature map by a block-wise transformation with secret key.\label{fig:transform}}
\end{figure*}

We utilize a block-wise transformation, Pixel Shuffling from~\cite{maungmaung2021protection}.
In this paper, a feature map $x$ in a dimension of $(c \times h \times w)$, where $c$ is the number of filters, $h$ is the height, and $w$ is the width of the feature map, is transformed with key $K$.
There are four steps in the process of transforming a feature map as shown in Fig.~\ref{fig:transform}.

\begin{enumerate}
  \item \textbf{Segment into blocks:} Segment $x$ into blocks with a size of $M$ such that $\{B_{(1,1)}, \ldots, B_{(\frac{h}{M}, \frac{w}{M})}\}$.
  \item \textbf{Flatten blocks:} Flatten each block $B_{(i, j)}$ into a vector $b_{(i,j)} = [b_{(i,j)}(1), \ldots, b_{(i,j)}(c \times M \times M)]$.
  \item \textbf{Transform blocks:} First, generate a random permutation vector $v = [v_1, \dots, v_k, \dots, v_{k'}, \dots, v_{c \times M \times M}]$ with key $K$, where $v_k \neq v_{k'}$ if $k \neq k'$, and shuffle every vector $b_{(i, j)}$ with $v$ as
    \begin{equation}
    b'_{(i, j)}(k) = b_{(i, j)}(v_k),
    \end{equation}
   to obtain a shuffled vector, \\$b'_{(i, j)} = [b'_{(i,j)}(1), \dots, b'_{(i,j)}(c \times M \times M)]$.
 \item \textbf{Concatenate blocks:} Concatenate the shuffled vectors to form a transformed feature map $x'$ in the dimension of $(c \times h \times w)$.
\end{enumerate}

If more than one feature map are required to be transformed, $K$ is applied to each feature map, and the above same process is repetitively carried out for all desired feature maps.

\subsection{Key Space}
In the previous method~\cite{maungmaung2021protection}, key space $\mathcal{K}$ of key $K$ depends on the number of pixels in a block, where $c$ is fixed to 3 because there are $3$ channels in an input RGB color image.
In contrast, in a feature map, $c \in \{64, 128, 256, 512\}$ is the number of filters.
Therefore, the key space is substantially increased and is defined as
\begin{equation}
  \mathcal{K}(c \times M \times M) = (c \times M \times M)!,
\end{equation}
which is larger than the key space in the previous method~\cite{maungmaung2021protection}.

\subsection{Threat Model}
A threat model includes a set of assumptions such as an attacker's goals, knowledge, and capabilities.
An attacker may steal a model to achieve different goals.
In this paper, we consider the attacker's goal is to be able to make use of a stolen model.
Therefore, we consider the attacker may estimate a key or fine-tune the model in order to remove the key.
We assume the attacker obtains a clone of the model and a small subset of the training dataset.
We carry out the following possible attacks with the intent of stealing a model to evaluate the robustness of the proposed model protection method.
In experiments, the proposed method will be demonstrated to be robust against attacks.

\subsubsection{Key Estimation Attack\label{sec:key}}
In reality, the key estimation attack is hard to carry out for the proposed model protection because the location of the transformed feature map cannot be known from the model itself.
There are many layers in a conventional CNN architecture and brute-forcing all layers to estimate the key will not be feasible.
To be practical, the cost of an attack should always be lower than that of training a new model.
We consider a worst-case scenario in which an attacker obtains additional information about the transformed feature map and estimate the key heuristically by observing the accuracy over their test images.

Algorithm~\ref{algo:key-estimation} describes the process of estimating a key.
First, we randomly initialize a key $K'$.
Next, we also initialize a set of index pairs $\mathcal{P}$ as $\mathcal{P} = \{(1, 2), (1, 3), \ldots, (c \times M \times M - 1, c \times M \times M)\}$ for a random permutation vector, $v'$ generated by $K'$. The number of all possible combinations of pairs for $v'$ can be computed as a binomial coefficient given by
\begin{equation}
  \left| \mathcal{P} \right| = {}_n C_r = \frac{n!}{r!(n-r)!},
\end{equation}
where $n = c \times M \times M$, and $r = 2$. For each index pair, we swap the pair in $v'$ if the swap improves the accuracy as shown in Algorithm~\ref{algo:key-estimation}.

Key estimation attacks do not guarantee that the attacker will find the correct key because the attacker does not know the actual performance of the correct key. However, the attacker may perform fine-tuning attacks to exploit a stolen model as below.

\begin{algorithm}
  \caption{Key Estimation\label{algo:key-estimation}}
\begin{algorithmic}[1]
\renewcommand{\algorithmicrequire}{\textbf{Input:}}
\renewcommand{\algorithmicensure}{\textbf{Output:}}
\REQUIRE{Input images with labels}
\ENSURE{$v'$}
\STATE{Initialize $K'$ randomly.}
\STATE{Initialize $\mathcal{P} = \{(1, 2), (1, 3), \ldots, (i, j), \ldots, (c \times M \times M - 1, c \times M \times M)\}$}
\STATE{Generate $v'$ by $K'$.}
\STATE{accuracy $\leftarrow$ Calculate accuracy of input images}
\FOR{Each index pair $(i, j)$ in $\mathcal{P}$}
\STATE{$(v'_{i}, v'_{j}) \leftarrow (v'_{j}, v'_{i})$}
\STATE{current\_accuracy $\leftarrow$ Calculate accuracy of input images}
\IF{current\_accuracy $<$ accuracy}
\STATE{// Accuracy does not improve, return the swap.}
\STATE{$(v'_{i}, v'_{j}) \leftarrow (v'_{j}, v'_{i})$}
\ELSE
\STATE{accuracy $\leftarrow$ current\_accuracy}
\ENDIF
\ENDFOR
\end{algorithmic}
\end{algorithm}

\subsubsection{Fine-tuning Attack}
In practice, CNNs are not trained from the beginning with random weights because creating a large dataset like ImageNet is difficult and expensive. Therefore, CNNs are usually pre-trained with a larger dataset (e.g., ImageNet), known as transfer learning~\cite{2015-ICLR-Simonyan}, which is to train a model on top of pre-trained weights.
An attacker may use fine-tuning to remove a key from a protected model.
We can consider such an attack scenario where the adversary has a subset of dataset $D'$ and fine-tunes the model.

\robustify\bfseries
\sisetup{table-parse-only,detect-weight=true,detect-inline-weight=text,round-mode=places,round-precision=2}
\begin{table*}[tbp]
\centering
\caption{Accuracy (\SI{}{\percent})  and key space of proposed protected models comparing with previous protected ones and baseline model\label{tab:results}}
\begin{tabular}{clSSSS}

  \toprule
  & {Model} & {Key Space} & {Accuracy ($K$)} & {Accuracy ($K'$)} & {Accuracy (without transformation)}\\
  \midrule
  \multirow{5}{*}{\rot{Proposed}} & {Initial Conv ($M = 2$)} & 256! & 94.83 & 10.74 & 9.94\\
                                  & {Layer 1 ($M = 2$)} & 256! & 95.38 & 9.64 & 10.08\\
                                  & {Layer 2 ($M = 2$)} & 512! & 95.16 & 10.64 & 6.55\\
                                  & {Layer 3 ($M = 2$)} & 1024! & 95.39 & 10.16 & 10.22\\
                                  & {Layer 4 ($M = 2$)} & 2048! & 95.21 & 11.36 & 1.30\\
  \midrule
  \multirow{6}{*}{\rot{Previous\cite{maungmaung2021protection}}} & {SHF ($M = 2$)} & {12!} & 94.76 & 36.55 & 31.43\\
                                                              & {NP ($M = 2$)} & {$2^{12}$} & 95.32 & 19.40 & 13.91\\
                                                              & {FFX ($M = 2$)} & {$2^{12}$} & 93.80 & 14.67 & 38.84\\
                                                              & {SHF ($M = 4$)} & {48!} & 92.58 & 20.15 & 27.77\\
                                                              & {NP ($M = 4$)} & {$2^{48}$} & 93.41 & 12.50 & 12.17\\
                                                              & {FFX ($M = 4$)} & {$2^{48}$} & 92.29 & 18.45 & 37.06\\
  \midrule
                                                              & {Baseline} & \multicolumn{4}{c}{95.45 (Not protected)}\\
  \bottomrule
\end{tabular}
\end{table*}

\section{Experiment Results}

\subsection{Setup\label{sec:setup}}
We conducted image classification experiments on the CIFAR-10 dataset~\cite{2009-Report-Krizhevsky} with a batch size of 128 and live augmentation (random cropping with padding of 4 and random horizontal flip) on a training set.
CIFAR-10 consists of 60,000 color images (dimension of $32 \times 32 \times 3$) with 10 classes (6000 images for each class) where 50,000 images are for training and 10,000 for testing.
We used deep residual networks~\cite{2016-CVPR-He} with 18 layers (ResNet-18) and trained for $200$ epochs with cyclic learning rates~\cite{2017-Arxiv-Smith} and mixed precision training~\cite{2017-Arxiv-Micikevicius}.
The parameters of the stochastic gradient descent (SGD) optimizer were: momentum of $0.9$, weight decay of $0.0005$ and maximum learning rate of $0.2$.

\subsection{Classification Performance}
We trained five protected models by applying a block-wise transformation with a secret key to different feature maps in ResNet-18.
All models in the experiments used a block size $M$ of 2.
We tested the protected models under three conditions for transformation: with correct key $K$, with incorrect key $K'$, and without applying the transformation.

Table~\ref{tab:results} summarizes the results of proposed protected models comparing with the previous protected models~\cite{maungmaung2021protection}, where the classification accuracy for incorrect key $K'$ was averaged over 100 random keys.
The key space for all models is also presented in Table~\ref{tab:results}.
The model trained by transforming the feature map of the initial convolution is indicated as ``Initial Conv'', that of the first group of residual blocks as ``Layer 1'', the second as ``Layer 2'', and so on (see also Fig.~\ref{fig:protection2}).
The previous protected models are named after the shorthand of the type of transformation; the model trained by using images transformed by Pixel Shuffling is denoted as SHF, that by negative/positive transformation as NP, and that by Feistel-based Format Preserving Encryption as FFX in Table~\ref{tab:results}.
Experiment results show that the accuracy of the proposed models are almost the same as that of non-protected accuracy (i.e., Baseline).
Moreover, the proposed models significantly increased the key space, and maintained a higher classification accuracy for correct key $K$, a lower classification accuracy for incorrect key $K'$ and without transformation.
Therefore, the proposed models outperformed the previous protected models in any case.

\subsection{Robustness Against Key Estimation Attack}
The proposed method was evaluated against key estimation attack in accordance with Algorithm~\ref{algo:key-estimation}.
As described in Section~\ref{sec:key}, elements in $v'$ which are generated by $K'$ were rearranged in accordance with the improvement in accuracy, and the resulting estimated $v'$ was used to evaluate the performance of the protected models.

Table~\ref{tab:esti} captures the classification performance of the proposed protected models comparing with the previous protected ones under the key estimation attack.
Note that we compared the previous models with $M = 4$ because the key space of the previous models for $M = 2$ is relatively small.
We observed that the accuracy of estimated key $K'$ for Layer 3 and 4 are \SI{50.13}{\percent} and \SI{89.30}{\percent} respectively.
Interestingly, although the key space of Layer 3 and 4 was larger, key estimation attacks found a good key to provide a reasonable accuracy.
However, the estimated keys were not good enough to provide a reasonable accuracy for the other models.
Comparing with the previous protected models, the proposed models provided better resistance against key estimation attacks except the models: Layer 3 and 4.

\robustify\bfseries
\sisetup{table-parse-only,detect-weight=true,detect-inline-weight=text,round-mode=places,round-precision=2}
\begin{table}[tbp]
\centering
\caption{Accuracy (\SI{}{\percent})  of proposed models under key estimation attack comparing with previous protected models\label{tab:esti}}
\begin{tabular}{lSS}

  \toprule
  {Model} & {Correct ($K$)} & {Estimated ($K'$)}\\
  \midrule
  {Initial Conv ($M = 2$)} & 94.83 & 20.17\\
  {Layer 1 ($M = 2$)} & 95.38 & 16.35\\
  {Layer 2 ($M = 2$)} & 95.16 & 23.58\\
  {Layer 3 ($M = 2$)} & 95.39 & 50.13\\
  {Layer 4 ($M = 2$)} & 95.21 & 89.30\\
  \midrule
  {SHF ($M = 4$)\cite{maungmaung2021protection}} & 92.58 & 25.66\\
  {NP ($M = 4$)\cite{maungmaung2021protection}} & 93.41 & 37.44\\
  {FFX ($M = 4$)\cite{maungmaung2021protection}} & 92.29 & 80.97\\
  \bottomrule
\end{tabular}
\end{table}

\subsection{Robustness Against Fine-tuning Attack}
We ran an experiment with different sizes of the attacker's dataset (i.e., $\left| D' \right| \in \{100, 500, 1000\}$).
We fine-tuned the models with $D'$ for 30 epochs with the same training settings in Section~\ref{sec:setup}.
Table~\ref{tab:fine-tune} shows the results of fine-tuning attacks for the proposed protected models comparing with the previous protected models.
Although the accuracy improved with respect to the size of the adversary's dataset, it was still lower than the performance of the correct key $K$ except for Layer 3 and 4.
Comparing to the previous protected models, the model ``Initial Conv'' provided better robustness against fine-tuning attacks than any other models.

\robustify\bfseries
\sisetup{table-parse-only,detect-weight=true,detect-inline-weight=text,round-mode=places,round-precision=2}
\begin{table}[tbp]
\centering
\caption{Accuracy (\SI{}{\percent})  of proposed models under fine-tuning attacks comparing with previous protected models. All models are with $M = 2$.\label{tab:fine-tune}}
\resizebox{\columnwidth}{!}{%
\begin{tabular}{lSSSS}

  \toprule
  {Model} & {Original} & {$\left| D' \right| = 100$} & {$\left| D' \right| = 500$} & {$\left| D' \right| = 1000$}\\
  \midrule
  {Initial$^{\dagger}$} & 94.83 & 18.47 & 55.38 & 69.42\\
  {Layer 1} & 95.38 & 22.37 & 66.90 & 78.54\\
  {Layer 2} & 95.16 & 21.75 & 66.38 & 74.94\\
  {Layer 3} & 95.39 & 20.84 & 74.59 & 80.99\\
  {Layer 4} & 95.21 & 87.43 & 73.28 & 94.63\\
  \midrule
  {SHF\cite{maungmaung2021protection}} & 94.76 & 33.89 & 70.69 & 78.01\\
  {NP\cite{maungmaung2021protection}} & 95.32 & 16.84 & 58.17 & 75.00\\
  {FFX\cite{maungmaung2021protection}} & 93.80 & 44.32 & 75.45 & 80.86\\
  \bottomrule
  \multicolumn{5}{l}{$^{\dagger}$Initial stands for ``Initial Conv''.}
\end{tabular}
}
\end{table}

\subsection{Comparison with State-of-the-art DNN Access Control Method}
Since underlying mechanisms of the DNN access control method by~\cite{chen2018protect} which uses a perturbation network and the proposed model protection method are different, it is difficult to directly compare them.
To make a high-level comparison, we implemented the anti-piracy method~\cite{chen2018protect} in the same training settings as in Section~\ref{sec:setup} for the CIFAR-10 dataset.
Then, we compared the anti-piracy method~\cite{chen2018protect} with the proposed model protection method (``Initial Conv'' model) in terms of authorized accuracy (i.e., with correct transformation/perturbation), unauthorized accuracy (i.e., without transformation/perturbation), and the core method used in the two mechanisms (Table~\ref{tab:comparison}).
The main difference is that the proposed model protection method uses a block-wise transformation with a secret key and the anti-piracy method~\cite{chen2018protect} utilizes a perturbation network.
In terms of classification performance, the proposed method achieves a higher authorized accuracy, which is close to baseline accuracy, and a lower unauthorized accuracy than that of the anti-piracy method~\cite{chen2018protect}.

\robustify\bfseries
\sisetup{table-parse-only,detect-weight=true,detect-inline-weight=text,round-mode=places,round-precision=2}
\begin{table}[tbp]
\centering
\caption{Comparison of proposed protected model and state-of-the-art anti-piracy model~\cite{chen2018protect}\label{tab:comparison}}
\resizebox{\columnwidth}{!}{%
\begin{tabular}{lSSl}

  \toprule
  {Model} & {Authorized} & {Unauthorized} & {Method}\\
  & {Accuracy} & {Accuracy} & \\
  \midrule
  {Initial Conv (Proposed)} & 94.83 & 9.94 & Block-Wise Transformation\\
  {Anti-piracy~\cite{chen2018protect}} & 92.89 & 14.21 & Perturbation Network\\
  \midrule
  {Baseline} & 95.45 & 95.45 & Non-protected\\
  \bottomrule
\end{tabular}
}
\end{table}

\section{Conclusion}
We proposed a model protection method by directly applying a block-wise transformation with a secret key to feature maps in the network.
As a result, the proposed model protection method not only improves the classification accuracy, but also increases the key space substantially.
The performance accuracy of the proposed protected model was closer to that of a non-protected model when the key was correct, and it dropped drastically when an incorrect key was given, suggesting that the proposed model is not usable to its full capacity for unauthorized users.
Experiments results show that the proposed models outperformed the previous protected models in terms of classification accuracy and robustness against key estimation attacks and fine-tuning attacks.

\bibliographystyle{IEEEtran}
\bibliography{IEEEabrv,refs}
\end{document}